\documentclass[runningheads]{llncs}
\usepackage[T1]{fontenc}
\usepackage{graphicx}
\usepackage{booktabs}
\usepackage{xcolor}
\usepackage{graphicx}
\usepackage{algorithm}
\usepackage{algorithmic}
\usepackage[misc]{ifsym}
\newcommand{\corr}{(\Letter)}
\usepackage{amsmath,amssymb,amsfonts}%
\usepackage{multirow}
\usepackage{mwe}
\usepackage{url}

\def\our{CROCE}

\begin{document}

\title{A Probabilistic Consensus-Driven Approach for Robust Counterfactual Explanations}

\titlerunning{A Probabilistic Approach for Robust Counterfactual Explanations}


\author{Marcin Kostrzewa\inst{1} \corr \and
Maciej Zięba\inst{1,2}  \and
Jerzy Stefanowski\inst{3}}

\authorrunning{M. Kostrzewa et al.}

\institute{Wrocław University of Science and Technology, Wrocław 50-347, Poland \email{\{marcin.kostrzewa,maciej.zieba\}@pwr.edu.pl}
\and
Tooploox, Wrocław 53-601, Poland
\and
Poznań University of Technology, Poznań 61-131, Poland
\email{jerzy.stefanowski@cs.put.poznan.pl}}

\maketitle

\begin{abstract}

Counterfactual explanations (CFEs) are essential for interpreting black-box models, yet they often become invalid when models are slightly changed. Existing methods for generating robust CFEs are often limited to specific types of models, require costly tuning, or inflexible robustness controls. We propose a novel approach that jointly models the data distribution and the space of plausible model decisions to ensure robustness to model changes. Using a probabilistic consensus over a model ensemble, we train a conditional normalizing flow that captures the data density under varying levels of classifier agreement. At inference time, a single interpretable parameter controls the robustness level; it specifies the minimum fraction of models that should agree on the target class without retraining the generative model. Our method effectively pushes CFEs toward regions that are both plausible and stable across model changes. Experimental results demonstrate that our approach achieves superior empirical robustness while also maintaining good performance  across other evaluation measures.

\end{abstract}

\section{Introduction}






Counterfactual explanations (CFEs) have emerged as one of the most popular tools for interpreting decisions made by complex, black-box machine learning models \cite{Guidotti2022survey,Verma2024survey}. Given an instance classified unfavorably, a counterfactual explanation identifies the minimal changes to its features that would lead to a different, desired outcome. This form of explanation is naturally human-readable, actionable, and closely aligned with legal frameworks such as the GDPR's right to explanation \cite{Wachter2018method}. As a result, CFEs have attracted broad research interest and found applications in high-stakes domains, including credit scoring \cite{Wachter2018method}, medical diagnosis \cite{Mertes2022medical}, and many others \cite{Guidotti2022survey}.

A good counterfactual explanation is expected to satisfy several desirable properties beyond validity (i.e., leading to the desired prediction), including proximity to the original instance, sparsity of feature changes, plausibility with respect to the data distribution (i.e., it should be located inside a dense class distribution), actionability, and others. Because these properties can be formalized and traded off in many ways, numerous generation methods have been proposed \cite{Guidotti2022survey,Verma2024survey}. But nearly all of them address static scenarios.

\begin{figure}[h!]
    \centering
    \includegraphics[width=0.7\linewidth]{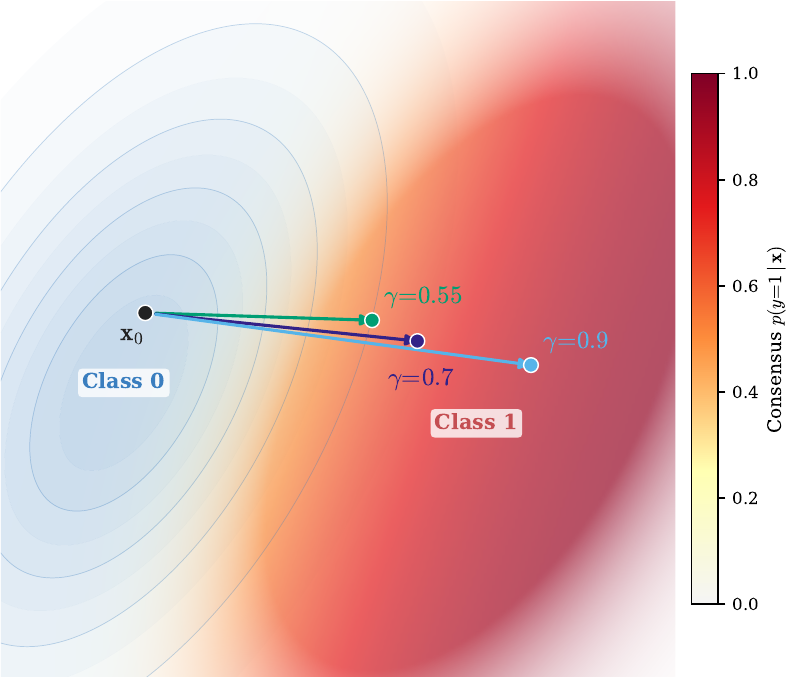}
    \caption{Illustration of the consensus-driven mechanism on a synthetic dataset. The background color for Class~1 encodes the values of ensemble consensus $p(y = 1 | \mathbf{x})$. As the required agreement level $\gamma$ increases, the generated counterfactual moves away from the decision boundary into regions of higher classifier consensus, trading proximity for robustness.}
    \label{fig:blobs-consensus}
\end{figure}

However, in practice, CFEs face an additional challenge that static evaluation does not capture: i.e., the underlying model may slightly change between the time a counterfactual is issued as a recommendation and the time a user acts upon it. It may potentially invalidate this previously generated CFE and make the user's efforts to fulfill these requirements  ineffective. In such situations, it is rather necessary to maintain the validity of the proposed explanation even after such a change.

To illustrate the importance of this problem, let us consider algorithmic decision-making institutions such as banks or insurers, where models are periodically retrained or updated as new data becomes available. Consider a loan applicant who receives a counterfactual recommendation to increase their savings by a certain amount; by the time they are able to do so, the institution may have slightly retrained its model, shifting the decision boundary such that the improved application is still rejected. This motivates the notion of robustness to model change \cite{Jiang2024survey}: the requirement that a CFE remains valid under sufficiently small changes of the underlying model.

Robustness to model change has recently gained  research attention \cite{Jiang2024survey}, and a number of methods have been proposed, broadly divided into end-to-end \cite{nguyen2022robust,Upadhyay2021Roar} and post-hoc approaches \cite{dutta2022robx,stepka2025betarce}. However, existing methods suffer from limitations such as being restricted to specific model families, requiring costly hyperparameter tuning, or lacking interpretable control over the robustness level. Importantly, most of these approaches address robustness purely through optimization adjustments or post-processing corrections, without considering how model predictions vary across the feature space under plausible model changes. As a result, they offer limited insight into where in the feature space CFEs are likely to remain valid when the model changes.


In this work, we take a probabilistic perspective on the problem of robust CFEs generation. Rather than perturbing a fixed classifier, we model the joint distribution of data and plausible model decisions by using a probabilistic consensus over an ensemble of models drawn from the space of admissible retrainings. This consensus signal serves as a conditioning variable for a normalizing flow that models the data distribution \cite{Rezende2015flow}. Critically, our approach is fully classifier-agnostic: the normalizing flow is trained once and encodes robustness information in a way that is independent of any particular model architecture. At inference time, a single interpretable parameter $\gamma$, the minimum ensemble mean probability for the target class, controls the desired robustness level without any retraining. Figure~\ref{fig:blobs-consensus} illustrates this mechanism -- increasing $\gamma$ pushes the counterfactual further from the original instance but places it in a region where more models agree on the target class, improving robustness.


We make the following contributions:
\begin{itemize}
    \item[\textbullet] We propose a novel probabilistic framework for robust CFEs generation that jointly models the data distribution and the space of plausible model decisions through a consensus signal derived from an ensemble of models sampled from the admissible model space.
    \item[\textbullet] We introduce a classifier-agnostic generative model for robust CFEs based on a conditional normalizing flow, trained once offline, in which a single interpretable parameter $\gamma$, the minimum ensemble mean probability for the target class, controls the desired robustness level at inference time without any retraining.
    \item[\textbullet] We empirically demonstrate that our approach achieves superior robustness to model change while maintaining competitive performance across other evaluation metrics.
\end{itemize}

\section{Related Work}

Although many approaches for generating CFEs have already been proposed \cite{Guidotti2022survey}, most of them do not answer the question of whether the generated explanations remain valid when conditions change. The issue of robust counterfactual explanations has only recently received systematic attention. Following the taxonomy of \cite{Jiang2024survey}, robustness can be considered from several perspectives: against input changes (requiring that similar inputs receive consistent explanations), against imperfect execution (requiring that a CFE remains valid when the user slightly fails to achieve the prescribed feature values), against model multiplicity (concerning validity across subsets of an ensemble), and against model change (ensuring validity after the underlying model is slightly modified). In this work, we focus on robustness against model change, which is the most extensively studied variant and is relevant in practice, as models in deployment are routinely retrained on new data or with updated configurations. The three main categories of methods proposed for this type of robustness are described below.

\subsubsection{End-to-end methods.}
Several approaches incorporate robustness directly into the CFEs generation objective. ROAR \cite{Upadhyay2021Roar} optimizes a CFE to remain valid under the worst-case small perturbation of the decision boundary. However, for non-linear models, it relies on a LIME surrogate to approximate the classifier locally, which can introduce approximation errors. RBR \cite{nguyen2022robust} casts the problem as posterior probability ratio minimization and models the data distribution with Gaussian kernels to account for data perturbations, but the kernel density estimation scales poorly to higher dimensions. Recent empirical comparisons \cite{stepka2025betarce} have shown that both methods achieve poor robustness in practice. \cite{Ferrario2022robust} takes a different route by augmenting the training set with previously issued CFEs, which fundamentally alters the model's training procedure and is inapplicable when the model is managed externally.

\subsubsection{Post-hoc methods.}
RobX \cite{dutta2022robx} defines a counterfactual stability metric that estimates local class variability around a CFE by querying the model on synthetic neighbors and iteratively refines a base CFE until a local class stability threshold is met. While model-agnostic, it assumes a well-calibrated model and requires careful dataset-specific tuning of its hyperparameters, and its stability metric lacks an intuitive interpretation in terms of the likelihood of remaining valid under model change. BetaRCE \cite{stepka2025betarce} grounds robustness in a following probabilistic framework: it defines a space of admissible models and estimates the probability that a CFE remains valid under a randomly drawn model from this space, accepting candidates that satisfy a $(\delta, \alpha)$-robustness condition. BetaRCE is classifier-agnostic and provides interpretable probabilistic guarantees, but its ensemble must be queried at every candidate evaluation during the search, making it computationally expensive per instance.

\subsubsection{Methods with formal guarantees.}

There are also other methods besides BetaRCE that offer theoretical guarantees, e.g., \cite{jiang2024probvablyrobust,jiang2024interval,marzari2024rigorousprobabilisticguaranteesrobust}. Despite providing strong theoretical grounding, all such methods are restricted to specific model families (primarily neural networks) and define model change exclusively as weight-space perturbation.

\subsubsection{Limitations of existing approaches.} 
To summarize, existing methods for robust counterfactual generation either rely on local model queries whose reliability depends on classifier calibration (RobX), require querying an ensemble at every candidate evaluation during search (BetaRCE), or are restricted to specific model families (methods with formal guarantees). Crucially, none of them explicitly model how the data distribution and classifier predictions jointly vary across the feature space under plausible model changes, nor do they offer an interpretable way of controlling the minimum level of robustness at inference time without retraining. Our approach addresses both of these gaps.

\section{Method}

In this section, we describe our approach for generating robust counterfactual explanations. We begin by recalling the standard counterfactual generation formulation and the basics of normalizing flows. We then formalize the robust counterfactual generation problem by defining a space of admissible models and introducing a probabilistic objective that requires the counterfactual to achieve high data density across all admissible levels of model agreement (consensus). Finally, we present our method: a conditional normalizing flow trained on consensus values derived from a model ensemble, which allows robust counterfactual generation with a single interpretable parameter controlling the robustness level at inference time, without retraining.

\subsection{Background}

\subsubsection{Counterfactual explanations generation.}
Following the framework established by Wachter et al.~\cite{Wachter2018method}, a counterfactual explanation identifies a perturbed instance $\mathbf{x}' \in \mathbb{R}^d$ derived from an initial point $\mathbf{x}_0 \in \mathbb{R}^d$. The primary objective is to shift the prediction of a classification model $M$ toward a target category $y'$, such that $M(\mathbf{x}') = y'$. Finding the optimal $\mathbf{x}'$ is typically framed as the following minimization problem:

\begin{equation}
    \arg\min_{\mathbf{x}' \in \mathbb{R}^d} d(\mathbf{x}_0, \mathbf{x}') + \lambda \cdot \ell_{v}(M(\mathbf{x}'), y'),
\end{equation}
where $\ell_{v}(\cdot, \cdot)$ is classification loss (usually cross-entropy), and $d(\cdot, \cdot)$ is distance measure, usually assumed to be $L_1$ (Manhattan) or $L_2$ (Euclidean) norm. Hyperparameter $\lambda$ controls the trade-off between the two loss components. 

Some recent works are focused on plausibility constraints that assume the example should be located in high-density regions \cite{Artelt2020plausibile,Wielopolski2024ppcef}. In such a scenario, the objective can be enriched with an additional component:

\begin{equation}
    \arg\min_{\mathbf{x}'  \in \mathbb{R}^d} d(\mathbf{x}_0, \mathbf{x}') + \lambda \cdot \biggl( \ell_{v}(M(\mathbf{x}'), y')+ \ell_{p}(\mathbf{x}', y') \biggr)  \text{,}
\label{eq:our_cf1}
\end{equation}
where $\ell_{p}(\cdot,\cdot)$ is the plausibility loss, usually represented by the conditional log-likelihood function.  

\subsubsection{Normalizing flows.}
Normalizing flows~\cite{Rezende2015flow} are generative models that learn an invertible, differentiable transformation $f_\theta \colon \mathbb{R}^d \to \mathbb{R}^d$ mapping a simple base distribution $p_z(\mathbf{z})$ (e.g., a standard Gaussian) to a complex target distribution $p_\theta(\mathbf{x})$. By the change-of-variables formula, the log-density of a point $\mathbf{x} = f_\theta(\mathbf{z})$ is:
\begin{equation}
    \log p_\theta(\mathbf{x}) = \log p_z\bigl(f_\theta^{-1}(\mathbf{x})\bigr) + \log \left| \det \frac{\partial f_\theta^{-1}}{\partial \mathbf{x}} \right|.
\end{equation}
Conditional normalizing flows extend this by parameterizing the transformation with a context variable $\mathbf{c}$, yielding a conditional density $p_\theta(\mathbf{x} \mid \mathbf{c})$. In practice, $\mathbf{c}$ is fed as an additional input to the neural networks defining each flow layer, so the learned transformation adapts to the conditioning value. This allows modeling distributions that vary smoothly with a continuous context signal, a property we exploit in our method.

\subsection{Robust counterfactual generation problem}
\label{sec:robust_problem}

The formulation~\eqref{eq:our_cf1} defines validity with respect to a single fixed model $M$. In practice, $M$ may be retrained before the user acts on the explanation, potentially invalidating it. We therefore seek counterfactuals that remain valid not for one model but across a distribution of plausible models.

Following \cite{stepka2025betarce}, we define the \emph{space of admissible models} $\mathcal{M}$ as the distribution over models obtained by retraining $M$ under permissible changes (i.e., different data samples, random seeds, or hyperparameters). For each model $M_i \in \mathcal{M}$, we can compute $p(y \mid \mathbf{x}, M_i)$, and the predictive distribution marginalized over $\mathcal{M}$ is:
\begin{equation}
    p(y \mid \mathbf{x}) = \int_{\mathcal{M}} p(M) \, p(y \mid \mathbf{x}, M) \, dM,
    \label{eq:marginal_predictive}
\end{equation}
approximated with $K$ models sampled uniformly from $\mathcal{M}$ as:
\begin{equation}
    p(y \mid \mathbf{x}) \approx \frac{1}{K} \sum_{k=1}^{K} p(y \mid \mathbf{x}, M_k).
    \label{eq:ensemble_approx}
\end{equation}

The quantity $p(y = c' \mid \mathbf{x})$ is the probability that a randomly drawn admissible model predicts class $c'$ at $\mathbf{x}$, a natural measure of how robust a point is with respect to model change. We parameterize the required minimum level of agreement by $\gamma \in (0,1]$. A robust counterfactual should be close to $\mathbf{x}_0$ and lie in a region where at least a fraction $\gamma$ of admissible models agree on class $c'$, while also being plausible with respect to the data distribution. Therefore, we propose using the conditional distribution over the data, where the conditioning represents the desired consensus level, denoted by $s$. In such a scenario, the objective can be formulated as:

\begin{equation}
    \arg\min_{\mathbf{x}'} \; d(\mathbf{x}_0, \mathbf{x}') - \alpha \cdot \log  \bigl[ p(\mathbf{x}' \mid s) \bigr],
    \label{eq:prerobust_objective}
\end{equation}
where $p(\mathbf{x}' \mid s)$ denotes the data density at the consensus level $s > \gamma$ and $\alpha$ controls the trade-off between proximity and plausibility. Thanks to this formulation, the plausibility and validity components from \eqref{eq:our_cf1} are represented by a single likelihood component. However, it raises two fundamental challenges: how to model the conditional distribution $p(\mathbf{x}' \mid s)$, and how to set the consensus level $s$. Higher values of $s$ push the counterfactual further from the decision boundary into regions of stronger model agreement, improving robustness but potentially increasing the distance from $\mathbf{x}_0$.

\subsection{Modeling $p(\mathbf{x}' \mid s)$ }

Following previous works \cite{Furman2025dicoflex,Wielopolski2024ppcef}, we propose to utilize normalizing flows for modeling the conditional distribution $p(\mathbf{x}' \mid s)$. We focus on binary classification and further elaborate on how it can be extended to multiclass scenarios.  In practice, the ensemble $M_1, \ldots, M_K$ can be constructed in various ways depending on the type of admissible model changes one wishes to capture, e.g., by bootstrap resampling of the training data, varying random seeds, or introducing small perturbations to hyperparameters. For each training point $\mathbf{x}_i$, we compute the \emph{consensus value} $s(\mathbf{x}_i) = \frac{1}{K}\sum_{k} p(y = 1 \mid \mathbf{x}_i, M_k) \in [0,1]$, the ensemble's mean probability of class~1. This single scalar encodes agreement for both classes: high $s$ indicates consensus for class~1, and low $s$ indicates consensus for class~0.   

We train a conditional normalizing flow $p_\theta(\mathbf{x} \mid s)$ on pairs $\{(\mathbf{x}_i, s(\mathbf{x}_i))\}_{i=1}^{N}$ by maximizing $\sum_i \log p_\theta(\mathbf{x}_i \mid s(\mathbf{x}_i))$. The resulting model captures the data density at every level of classifier agreement: $p_\theta(\mathbf{x} \mid s)$ describes where data points lie when the ensemble consensus is equal to $s$. Since the ensemble models serve only to produce the consensus values, they need not be differentiable, and thus the framework is fully \emph{classifier-agnostic}.

\begin{algorithm}[!t]
\caption{Proposed method}
\label{alg:rpcef}
\textbf{Offline:}
\begin{algorithmic}[1]
\REQUIRE data $\mathcal{D}$, ensemble $M_1,\ldots,M_K \sim \mathcal{M}$
\STATE Compute $s(\mathbf{x}_i) \leftarrow \frac{1}{K}\sum_k p(y{=}1 \mid \mathbf{x}_i, M_k)$ for all $i$
\STATE Train flow $p_\theta$ on $\{(\mathbf{x}_i, s(\mathbf{x}_i))\}$
\end{algorithmic}
\vspace{0.2em}
\textbf{Inference:} given $\mathbf{x}_0$, target $c'$, robustness $\gamma$
\begin{algorithmic}[1]
\STATE Initialize $\boldsymbol{\delta} \leftarrow \mathbf{0}$,\; $s \leftarrow \gamma$ \hfill (or $s \leftarrow 1-\gamma$ if $c' = 0$)
\FOR{$t = 1,\ldots,T$}
    \STATE $\mathcal{L} \leftarrow \|\boldsymbol{\delta}\|_1 + \alpha[\tau - \log p_\theta(\mathbf{x}_0 + \boldsymbol{\delta} \mid s)]^+$
    \STATE $\boldsymbol{\delta} \leftarrow \boldsymbol{\delta} - \eta \nabla_{\boldsymbol{\delta}}\mathcal{L}$
    \STATE $s \leftarrow \mathrm{clamp}(s - \eta \nabla_{s}\mathcal{L},\; \gamma,\; 1)$ \hfill (or $\mathrm{clamp}(\cdot,\; 0,\; 1{-}\gamma)$ if $c' = 0$)
\ENDFOR
\RETURN $\mathbf{x}' = \mathbf{x}_0 + \boldsymbol{\delta}$
\end{algorithmic}
\end{algorithm}

\subsection{Counterfactual  explanation procedure}

The objective given by eq. \eqref{eq:prerobust_objective} requires setting the value of $s$ in order to solve the optimization problem. One approach is to sample multiple values from the $[\gamma,1]$ interval, which is computationally expensive. We therefore treat $s$ as an additional optimization variable constrained to $[\gamma, 1]$ and optimize jointly over the perturbation $\boldsymbol{\delta} = \mathbf{x}' - \mathbf{x}_0$ and the consensus level $s$. This can be understood as searching for the most favorable consensus level within the admissible range for each counterfactual. Additionally, rather than maximizing the log-likelihood without bound, we impose a plausibility threshold $\tau$, set to the median log-likelihood of the training data under the flow, following \cite{Artelt2020plausibile,Wielopolski2024ppcef}. The resulting hinge term penalizes counterfactuals that fall below this threshold but incurs no cost once plausibility is satisfied, allowing the $L_1$ term to drive the solution toward proximity. The final objective is defined as:
\begin{equation}
    \min_{\boldsymbol{\delta},\; s \in [\gamma, 1]} \; \|\boldsymbol{\delta}\|_1 + \alpha \cdot \bigl[\tau - \log p_\theta(\mathbf{x}_0 + \boldsymbol{\delta} \mid s)\bigr]^{+},
    \label{eq:cf_objective}
\end{equation}
where $[\cdot]^+ = \max(0,\cdot)$ and $\alpha$ balances proximity against the consensus-conditioned density. The constraint $s \geq \gamma$ ensures robustness, the flow density ensures plausibility, and the $L_1$ term ensures proximity. When targeting class~0, the constraint becomes $s \in [0, 1-\gamma]$.

Since $\gamma$ enters only as a bound on $s$, the flow is trained once and reused at any robustness level without retraining. The consensus conditioning also provides implicit validity: for a well-trained flow, high-density regions conditioned on $s \geq \gamma$ correspond to areas of the feature space where the ensemble agrees on the target class, steering the counterfactual toward the desired prediction without requiring an explicit classifier loss $\ell_v$. We note that this implicit validity relies on the flow being well-calibrated and is not formally guaranteed. The proposed procedure is given in Algorithm~\ref{alg:rpcef}.

\paragraph{Multi-class extension.}
For $C > 2$ classes, the consensus becomes a vector $\mathbf{s}(\mathbf{x}) \in [0,1]^C$ with $s_c(\mathbf{x}) = \frac{1}{K}\sum_k [M_k(\mathbf{x})]_c$, and the flow is conditioned on this vector. The consensus constraint generalizes to $s_{c'} \geq \gamma$ for the target class component.

\section{Experiments}

In this section, we evaluate our method\footnote{Code available at \url{https://anonymous.4open.science/r/robust-cfe-5ED1/}}, which we will denote as \our{} (\textbf{C}onsensus-\textbf{RO}bust \textbf{C}ounterfactual \textbf{E}xplanations). We first describe the experimental setup, including datasets, classifiers, model ensemble construction, and evaluation metrics (Section~\ref{sec:setup}). We then present the main comparison results (Section~\ref{sec:results}) and analyze the sensitivity of our method to its key hyperparameters (Section~\ref{sec:hyperparams}).

\subsection{Experimental setup}
\label{sec:setup}

\subsubsection{Datasets and classifiers.}

We evaluate on three binary tabular classification datasets commonly used in the counterfactual explanation literature, i.e.: Diabetes, HELOC, and a synthetic two-dimensional Moons dataset. All features are continuous and normalized to $[0,1]$ using min-max scaling fitted on the training set. Dataset characteristics are summarized in Table~\ref{tab:datasets}.

\begin{table}[h]
\centering
\caption{Summary of datasets used in the experiments. Classifier accuracy is the mean test accuracy across 5 folds.}
\label{tab:datasets}
\begin{tabular}{lrrrrr}
\toprule
Dataset & Samples & Features & Classes & MLP Acc. & Linear Acc. \\
\midrule
Moons       & 1\,024  & 2  & 2 & 0.997 & 0.853 \\
Diabetes    & 768     & 8  & 2 & 0.779 & 0.752 \\
HELOC       & 10\,459 & 23 & 2 & 0.711 & 0.681 \\
\bottomrule
\end{tabular}
\end{table}

For each dataset, we perform 5-fold cross-validation. In each fold, the data is split into three disjoint subsets: a training set (40\%), a validation pool (40\%), and a test set (20\%). Split indices are fixed across all methods to ensure a fair comparison.

We consider two classifier architectures as the explained model: a multi-layer perceptron (MLP) and a logistic regression model (LR), both tuned via Bayesian optimization with Optuna~\cite{ozaki2025optunahub}. Details of the search spaces and selected configurations are provided in the supplementary materials and the accompanying code repository. The best-performing configuration from each fold serves as the \emph{base model} $M$, the classifier whose decisions are to be explained and against which validity is evaluated. Using two architecturally different classifiers allows us to assess whether the robustness improvements generalize across model families. For each classifier type, we construct three independent sets of model variants:

\begin{itemize}
    \item[\textbullet] a \emph{consensus ensemble} of 40 models\footnote{We use 40 admissible models as increasing this number does not significantly affect the classifiers' consensus estimate; this finding is consistent with \cite{stepka2025betarce} and our analysis in supplementary materials further confirms it.} drawn by bootstrap resampling from the training set, used to compute the consensus values $s(\mathbf{x}_i)$ for flow conditioning during training,
    \item[\textbullet] a \emph{retrain evaluation ensemble} of 30 models, each trained on the full training set combined with a bootstrap sample from the validation pool, simulating a scenario where additional data becomes available and the model is retrained,
    \item[\textbullet] a \emph{bootstrap evaluation ensemble} of 30 models, each trained on a bootstrap sample drawn from the validation pool alone, simulating a more challenging scenario where the model is retrained on a different subset of the available data.
\end{itemize}
Both evaluation ensembles reflect plausible model changes under a stable data distribution. The underlying generating process remains the same, but the training data varies across models. The bootstrap ensemble represents a harder robustness test, as its variants may differ more substantially from the base classifier due to the smaller and more variable training sets. Importantly, neither evaluation ensemble is used during training of our method, ensuring that the robustness assessment reflects generalization to genuinely unseen model variations.

\subsubsection{Compared methods.}

We compare our approach against three reference methods representing different strategies for counterfactual generation:
\begin{itemize}
    \item[\textbullet] \textbf{PPCEF}~\cite{Wielopolski2024ppcef}: a non-robust baseline that generates plausible counterfactuals using a class-conditional normalizing flow. It optimizes for proximity and plausibility but does not incorporate any robustness mechanisms. Including PPCEF allows us to isolate the effect of our consensus-based conditioning since our method builds on a similar generative architecture,
    \item[\textbullet] \textbf{RobX}~\cite{dutta2022robx}: the method requires hyperparameter tuning; we perform a grid search using a validation dataset, evaluating a local stability threshold  $\tau$ starting from $0.5$ and incrementing by $0.1$; for each $\tau$, we generate counterfactuals for a subset of validation ($N=100$) and retain only those values that achieve both full coverage (a counterfactual is found for each instance) and perfect validity. Then, we report two variants: \emph{RobX (default)}, which selects the lowest eligible $\tau$, yielding counterfactuals closest to the original instances, and \emph{RobX (robust)}, which selects the highest eligible $\tau$, pushing counterfactuals further from the decision boundary to favor robustness. Hyperparameter $\gamma$ is kept fixed and equal to $0.5$,
    \item[\textbullet] \textbf{BetaRCE}~\cite{stepka2025betarce}: we adopt the parameter configuration reported to yield the highest robustness in the original work ($\alpha = 0.95$, $\delta = 0.9$).
\end{itemize}

As both RobX and BetaRCE are post-hoc methods, we use GrowingSpheres as the base counterfactual generator for them following \cite{stepka2025betarce}. 

For our \our{} method, we use a MAF~\cite{papamakarios2017maf}, which has been empirically shown to outperform other normalizing flow architectures on tabular data~\cite{Furman2025dicoflex,Wielopolski2024ppcef}. The flow hyperparameters were also tuned using Optuna; details are provided in the supplementary materials. The counterfactual optimization uses $T = 2000$ gradient steps with a learning rate of $\eta = 10^{-2}$. We report results for several values of the consensus threshold $\gamma \in \{0.70, 0.80, 0.90\}$, also to study the trade-off between robustness and proximity. Throughout the experiments, we set $\alpha = 5$; the choice of this parameter is discussed in Section~\ref{sec:hyperparams}.

\subsubsection{Metrics.}
We assess the generated counterfactual explanations along four dimensions:
\begin{itemize}
    \item[\textbullet] \textbf{Validity}: the fraction of counterfactuals classified into the target class by the base model $M$.
    \item[\textbullet] \textbf{Proximity}: the $L_1$ and $L_2$ distances between the original instance and the counterfactual, measuring the magnitude of the required change.
    \item[\textbullet] \textbf{Plausibility}: the mean $L_2$ distance to the $k{=}10$ nearest neighbors in the training set belonging to the target class, quantifying how close the counterfactual lies to the observed data manifold.
    \item[\textbullet] \textbf{Robustness}: the fraction of evaluation ensemble models that predict the required target class on the counterfactual. We report this separately for both evaluation ensembles: Rob.{\scriptsize(ret)}, measured on the 30 models trained on the full training set augmented with bootstrap samples from the validation pool, and Rob.{\scriptsize(bs)}, measured on the 30 models trained on bootstrap samples from the validation pool alone. The latter represents a harder robustness test.
\end{itemize}

All metrics are computed per instance and reported as mean value $\pm$ standard deviation across 5 folds.

\subsection{Comparison results}
\label{sec:results}

Tables~\ref{tab:combined_mlp} and~\ref{tab:combined_linear} present the results for the MLP and LR classifiers, respectively. We discuss the key findings across the four evaluation dimensions.

\subsubsection{Robustness and proximity.}

Our method consistently achieves the highest robustness scores across all datasets and both classifier types. On the MLP classifier, \our{} with $\gamma = 0.9$ reaches near-perfect robustness on all three datasets under both evaluation protocols. In contrast, the baselines exhibit a clear gap. BetaRCE, despite setting $\delta = 0.9$, achieves lower robustness on Diabetes (Rob.{\scriptsize(ret)}$= 0.972$, Rob.{\scriptsize(bs)}$= 0.855$) and HELOC (Rob.{\scriptsize(ret)}$= 0.927$, Rob.{\scriptsize(bs)}$= 0.792$). RobX (default) performs inconsistently: while adequate on Moons (Rob.{\scriptsize(bs)}$= 0.994$), it drops on Diabetes (Rob.{\scriptsize(bs)}$= 0.770$) and HELOC (Rob.{\scriptsize(bs)}$= 0.786$). We attribute this to RobX's reliance on the counterfactual stability metric, which queries the base model's predicted probabilities in a local neighborhood; when the classifier is not well calibrated, these local estimates are less reliable. RobX (robust), which selects the highest eligible $\tau$, improves robustness but at a steep cost in proximity ($L_1 = 1.450$ on Diabetes, $3.346$ on HELOC).

The advantage of \our{} is also visible for the LR classifier. On HELOC, RobX (default) achieves only Rob.{\scriptsize(ret)}$= 0.431$, Rob.{\scriptsize(bs)}$= 0.433$, and even its robust version, despite reaching Rob.{\scriptsize(bs)}$= 0.996$, does so at an $L_1$ cost of $4.679$. On Diabetes, RobX (default) drops to Rob.{\scriptsize(bs)}$= 0.413$ while RobX (robust) reaches Rob.{\scriptsize(bs)}$= 0.738$ but at $L_1 = 1.673$; our mildest setting ($\gamma = 0.7$) achieves Rob.{\scriptsize(bs)}$= 0.963$ at a comparable $L_1 = 1.895$, and $\gamma = 0.8$ reaches Rob.{\scriptsize(bs)}$= 0.977$. BetaRCE occupies a middle ground on both datasets but never matches \our{}'s best robustness. This confirms that our consensus-based approach generalizes well across classifier families, while methods that depend on local model queries are sensitive to calibration quality. RobX (robust) can partially compensate by selecting a high $\tau$, but at a disproportionate cost in proximity.

\subsubsection{Validity.}
All methods achieve high validity across most settings. \our{} maintains near-perfect validity in the vast majority of configurations, despite not using an explicit classifier loss term in its objective. 
We hypothesize that the consensus conditioning provides implicit validity: maximizing the flow density at high consensus levels naturally places counterfactuals in regions where the models agree on the target class.

\subsubsection{Plausibility.}
\our{} consistently produces plausible counterfactuals for the MLP classifier, with scores comparable to or better than the baselines. \our{} matches PPCEF's plausibility on Moons ($0.035$ vs.\ $0.034$) and HELOC ($0.437$ for $\gamma=0.7$ vs.\ $0.461$), while substantially outperforming BetaRCE ($0.073$ on Moons, $0.671$ on HELOC). This is a direct consequence of the normalizing flow's density model, which steers counterfactuals toward high-density regions of the data distribution by construction. On the LR classifier, the picture is more nuanced: at moderate $\gamma$ values, \our{} achieves competitive plausibility (e.g., $0.588$ on HELOC at $\gamma = 0.7$, close to RobX's $0.590$), but plausibility degrades more noticeably at higher $\gamma$, as the more rigorous consensus constraint forces counterfactuals into narrower regions of the feature space.

\subsubsection{Computational cost.}
The generation time of \our{} is independent of $\gamma$. Unlike RobX and BetaRCE, which process instances sequentially, \our{}'s gradient-based optimization is fully parallelizable. This is visible on HELOC (500 test instances), where \our{} matches BetaRCE's wall-clock time; on the smaller Diabetes set (151 instances) the benefit is less pronounced. Details are provided in the supplementary materials.

\subsubsection{Summary.}
Overall, our method achieves the highest robustness scores across all datasets and classifier types while maintaining sufficient proximity, strong plausibility, and near-perfect validity. The single parameter $\gamma$ provides transparent control over this trade-off: a practitioner can choose a moderate $\gamma$ for a balanced solution or increase it when robustness is critical, without retraining the generative model. For instance, on HELOC with the MLP classifier, increasing $\gamma$ from $0.7$ to $0.9$ raises Rob.{\scriptsize(bs)} from $0.929$ to $0.977$ while $L_1$ grows from $2.195$ to $3.245$, illustrating how this trade-off can be navigated in practice.

\begin{table*}[ht]
\centering
\caption{Counterfactual explanation results for MLP classifier. Mean $\pm$ std across 5 folds.}
\label{tab:combined_mlp}
\resizebox{\textwidth}{!}{
\begin{tabular}{llcccccc}
\toprule
Dataset & Method & Validity $\uparrow$ & $L_1$ $\downarrow$ & $L_2$ $\downarrow$ & Plausibility $\downarrow$ & Rob.{\scriptsize(ret)} $\uparrow$ & Rob.{\scriptsize(bs)} $\uparrow$ \\
\midrule
\multirow{7}{*}{Moons} & PPCEF & $\mathbf{1.000 \pm 0.000}$ & $0.446 \pm 0.030$ & $0.349 \pm 0.026$ & $\mathbf{0.034 \pm 0.002}$ & $\mathbf{1.000 \pm 0.000}$ & $\mathbf{1.000 \pm 0.000}$ \\
 & RobX & $0.997 \pm 0.006$ & $0.409 \pm 0.011$ & $0.335 \pm 0.008$ & $0.042 \pm 0.005$ & $0.997 \pm 0.006$ & $0.994 \pm 0.007$ \\
 & RobX (robust) & $\mathbf{1.000 \pm 0.000}$ & $0.827 \pm 0.017$ & $0.655 \pm 0.013$ & $0.043 \pm 0.004$ & $\mathbf{1.000 \pm 0.000}$ & $\mathbf{1.000 \pm 0.000}$ \\
 & BetaRCE & $\mathbf{1.000 \pm 0.000}$ & $\mathbf{0.238 \pm 0.004}$ & $\mathbf{0.189 \pm 0.002}$ & $0.073 \pm 0.003$ & $0.965 \pm 0.020$ & $0.896 \pm 0.057$ \\
 & \our{} ($\gamma$=0.70) & $\mathbf{1.000 \pm 0.000}$ & $0.405 \pm 0.016$ & $0.323 \pm 0.013$ & $0.035 \pm 0.002$ & $\mathbf{1.000 \pm 0.000}$ & $\mathbf{1.000 \pm 0.000}$ \\
 & \our{} ($\gamma$=0.80) & $\mathbf{1.000 \pm 0.000}$ & $0.405 \pm 0.016$ & $0.323 \pm 0.013$ & $0.035 \pm 0.002$ & $\mathbf{1.000 \pm 0.000}$ & $\mathbf{1.000 \pm 0.000}$ \\
 & \our{} ($\gamma$=0.90) & $\mathbf{1.000 \pm 0.000}$ & $0.405 \pm 0.016$ & $0.324 \pm 0.013$ & $0.035 \pm 0.002$ & $\mathbf{1.000 \pm 0.000}$ & $\mathbf{1.000 \pm 0.000}$ \\
\midrule
\multirow{7}{*}{Diabetes} & PPCEF & $0.999 \pm 0.003$ & $0.705 \pm 0.028$ & $0.300 \pm 0.014$ & $\mathbf{0.260 \pm 0.014}$ & $0.870 \pm 0.045$ & $0.743 \pm 0.068$ \\
 & RobX & $0.873 \pm 0.255$ & $\mathbf{0.546 \pm 0.083}$ & $\mathbf{0.270 \pm 0.042}$ & $0.295 \pm 0.030$ & $0.855 \pm 0.245$ & $0.770 \pm 0.220$ \\
 & RobX (robust) & $\mathbf{1.000 \pm 0.000}$ & $1.450 \pm 0.206$ & $0.677 \pm 0.106$ & $0.399 \pm 0.050$ & $0.999 \pm 0.001$ & $0.926 \pm 0.137$ \\
 & BetaRCE & $\mathbf{1.000 \pm 0.000}$ & $0.622 \pm 0.060$ & $0.284 \pm 0.024$ & $0.328 \pm 0.014$ & $0.972 \pm 0.031$ & $0.855 \pm 0.054$ \\
 & \our{} ($\gamma$=0.70) & $0.997 \pm 0.005$ & $0.632 \pm 0.045$ & $0.343 \pm 0.021$ & $0.266 \pm 0.018$ & $0.979 \pm 0.030$ & $0.902 \pm 0.064$ \\
 & \our{} ($\gamma$=0.80) & $\mathbf{1.000 \pm 0.000}$ & $0.783 \pm 0.113$ & $0.415 \pm 0.047$ & $0.284 \pm 0.031$ & $0.997 \pm 0.003$ & $0.960 \pm 0.042$ \\
 & \our{} ($\gamma$=0.90) & $\mathbf{1.000 \pm 0.000}$ & $0.988 \pm 0.194$ & $0.509 \pm 0.068$ & $0.315 \pm 0.055$ & $\mathbf{1.000 \pm 0.000}$ & $\mathbf{0.994 \pm 0.009}$ \\
 \midrule
\multirow{7}{*}{HELOC} & PPCEF & $0.997 \pm 0.002$ & $1.792 \pm 0.105$ & $0.505 \pm 0.037$ & $0.461 \pm 0.019$ & $0.795 \pm 0.071$ & $0.654 \pm 0.082$ \\
 & RobX & $\mathbf{1.000 \pm 0.000}$ & $\mathbf{1.392 \pm 0.079}$ & $\mathbf{0.401 \pm 0.021}$ & $0.562 \pm 0.016$ & $0.922 \pm 0.058$ & $0.786 \pm 0.072$ \\
 & RobX (robust) & $\mathbf{1.000 \pm 0.000}$ & $3.346 \pm 0.535$ & $1.003 \pm 0.166$ & $0.534 \pm 0.021$ & $\mathbf{0.999 \pm 0.001}$ & $0.968 \pm 0.016$ \\
 & BetaRCE & $\mathbf{1.000 \pm 0.000}$ & $1.859 \pm 0.103$ & $0.509 \pm 0.029$ & $0.671 \pm 0.021$ & $0.927 \pm 0.056$ & $0.792 \pm 0.061$ \\
 & \our{} ($\gamma$=0.70) & $\mathbf{1.000 \pm 0.000}$ & $2.195 \pm 0.191$ & $0.717 \pm 0.063$ & $\mathbf{0.437 \pm 0.028}$ & $0.985 \pm 0.018$ & $0.929 \pm 0.051$ \\
 & \our{} ($\gamma$=0.80) & $\mathbf{1.000 \pm 0.000}$ & $2.555 \pm 0.237$ & $0.822 \pm 0.080$ & $0.460 \pm 0.034$ & $0.997 \pm 0.004$ & $0.966 \pm 0.038$ \\
 & \our{} ($\gamma$=0.90) & $\mathbf{1.000 \pm 0.000}$ & $3.245 \pm 0.276$ & $1.015 \pm 0.100$ & $0.535 \pm 0.040$ & $0.999 \pm 0.002$ & $\mathbf{0.977 \pm 0.030}$ \\
\bottomrule
\end{tabular}
}
\end{table*}

\begin{table*}[ht]
\centering
\caption{Counterfactual explanation results for logistic regression classifier. Mean $\pm$ std across 5 folds.}
\label{tab:combined_linear}
\resizebox{\textwidth}{!}{
\begin{tabular}{llcccccc}
\toprule
Dataset & Method & Validity $\uparrow$ & $L_1$ $\downarrow$ & $L_2$ $\downarrow$ & Plausibility $\downarrow$ & Rob.{\scriptsize(ret)} $\uparrow$ & Rob.{\scriptsize(bs)} $\uparrow$ \\
\midrule
\multirow{6}{*}{Moons} & PPCEF & $0.823 \pm 0.251$ & $0.518 \pm 0.063$ & $0.406 \pm 0.052$ & $\mathbf{0.034 \pm 0.002}$ & $0.726 \pm 0.187$ & $0.669 \pm 0.151$ \\
& RobX & $\mathbf{1.000 \pm 0.000}$ & $\mathbf{0.460 \pm 0.029}$ & $\mathbf{0.384 \pm 0.019}$ & $0.075 \pm 0.011$ & $0.831 \pm 0.161$ & $0.767 \pm 0.204$ \\
 & RobX (robust) & $\mathbf{1.000 \pm 0.000}$ & $0.777 \pm 0.190$ & $0.607 \pm 0.134$ & $0.060 \pm 0.014$ & $0.937 \pm 0.078$ & $0.841 \pm 0.139$ \\
 & BetaRCE & $\mathbf{1.000 \pm 0.000}$ & $0.642 \pm 0.240$ & $0.467 \pm 0.181$ & $0.072 \pm 0.008$ & $0.993 \pm 0.007$ & $0.947 \pm 0.029$ \\
 & \our{} ($\gamma$=0.70) & $\mathbf{1.000 \pm 0.000}$ & $1.084 \pm 0.000$ & $0.788 \pm 0.000$ & $0.101 \pm 0.000$ & $\mathbf{1.000 \pm 0.000}$ & $\mathbf{1.000 \pm 0.000}$ \\
 & \our{} ($\gamma$=0.80) & $\mathbf{1.000 \pm 0.000}$ & $1.087 \pm 0.000$ & $0.793 \pm 0.000$ & $0.112 \pm 0.000$ & $\mathbf{1.000 \pm 0.000}$ & $\mathbf{1.000 \pm 0.000}$ \\
 & \our{} ($\gamma$=0.90) & $\mathbf{1.000 \pm 0.000}$ & $1.088 \pm 0.000$ & $0.793 \pm 0.000$ & $0.117 \pm 0.000$ & $\mathbf{1.000 \pm 0.000}$ & $\mathbf{1.000 \pm 0.000}$ \\
\midrule
\multirow{6}{*}{Diabetes} & PPCEF & $0.948 \pm 0.101$ & $0.746 \pm 0.105$ & $0.311 \pm 0.049$ & $\mathbf{0.278 \pm 0.016}$ & $0.373 \pm 0.018$ & $0.386 \pm 0.020$ \\
& RobX & $\mathbf{1.000 \pm 0.000}$ & $\mathbf{0.584 \pm 0.032}$ & $\mathbf{0.286 \pm 0.017}$ & $0.320 \pm 0.018$ & $0.398 \pm 0.027$ & $0.413 \pm 0.030$ \\
 & RobX (robust) & $\mathbf{1.000 \pm 0.000}$ & $1.673 \pm 0.386$ & $0.765 \pm 0.174$ & $0.423 \pm 0.101$ & $0.926 \pm 0.091$ & $0.738 \pm 0.168$ \\
 & BetaRCE & $\mathbf{1.000 \pm 0.000}$ & $1.134 \pm 0.196$ & $0.486 \pm 0.074$ & $0.453 \pm 0.057$ & $0.946 \pm 0.092$ & $0.797 \pm 0.179$ \\
 & \our{} ($\gamma$=0.70) & $0.997 \pm 0.005$ & $1.895 \pm 0.202$ & $0.820 \pm 0.108$ & $0.467 \pm 0.044$ & $\mathbf{1.000 \pm 0.000}$ & $0.963 \pm 0.035$ \\
 & \our{} ($\gamma$=0.80) & $0.997 \pm 0.005$ & $2.407 \pm 0.343$ & $1.041 \pm 0.170$ & $0.635 \pm 0.095$ & $\mathbf{1.000 \pm 0.000}$ & $\mathbf{0.977 \pm 0.020}$ \\
 & \our{} ($\gamma$=0.90) & $0.988 \pm 0.023$ & $2.473 \pm 0.392$ & $1.067 \pm 0.185$ & $0.671 \pm 0.109$ & $\mathbf{1.000 \pm 0.000}$ & $0.975 \pm 0.021$ \\
\midrule
\multirow{7}{*}{HELOC} & PPCEF & $0.996 \pm 0.004$ & $1.799 \pm 0.110$ & $0.505 \pm 0.038$ & $\mathbf{0.492 \pm 0.017}$ & $0.423 \pm 0.019$ & $0.417 \pm 0.017$ \\
& RobX & $\mathbf{1.000 \pm 0.000}$ & $\mathbf{1.299 \pm 0.029}$ & $\mathbf{0.377 \pm 0.009}$ & $0.590 \pm 0.016$ & $0.431 \pm 0.020$ & $0.433 \pm 0.020$ \\
 & RobX (robust) & $\mathbf{1.000 \pm 0.000}$ & $4.679 \pm 0.232$ & $1.349 \pm 0.070$ & $0.620 \pm 0.017$ & $\mathbf{1.000 \pm 0.000}$ & $0.996 \pm 0.008$ \\
 & BetaRCE & $\mathbf{1.000 \pm 0.000}$ & $2.704 \pm 0.325$ & $0.743 \pm 0.090$ & $0.795 \pm 0.035$ & $0.985 \pm 0.009$ & $0.903 \pm 0.046$ \\
 & \our{} ($\gamma$=0.70) & $\mathbf{1.000 \pm 0.000}$ & $3.376 \pm 0.525$ & $0.977 \pm 0.111$ & $0.588 \pm 0.067$ & $\mathbf{1.000 \pm 0.000}$ & $0.999 \pm 0.001$ \\
 & \our{} ($\gamma$=0.80) & $0.998 \pm 0.005$ & $4.413 \pm 0.758$ & $1.229 \pm 0.167$ & $0.761 \pm 0.139$ & $\mathbf{1.000 \pm 0.000}$ & $\mathbf{1.000 \pm 0.000}$ \\
 & \our{} ($\gamma$=0.90) & $0.988 \pm 0.021$ & $4.899 \pm 0.367$ & $1.343 \pm 0.076$ & $0.871 \pm 0.091$ & $0.998 \pm 0.004$ & $0.998 \pm 0.003$ \\
\bottomrule
\end{tabular}
}
\end{table*}

\subsection{Hyperparameters analysis}
\label{sec:hyperparams}

\begin{figure}[h!]
    \centering
    \includegraphics[width=12cm, height=6cm]{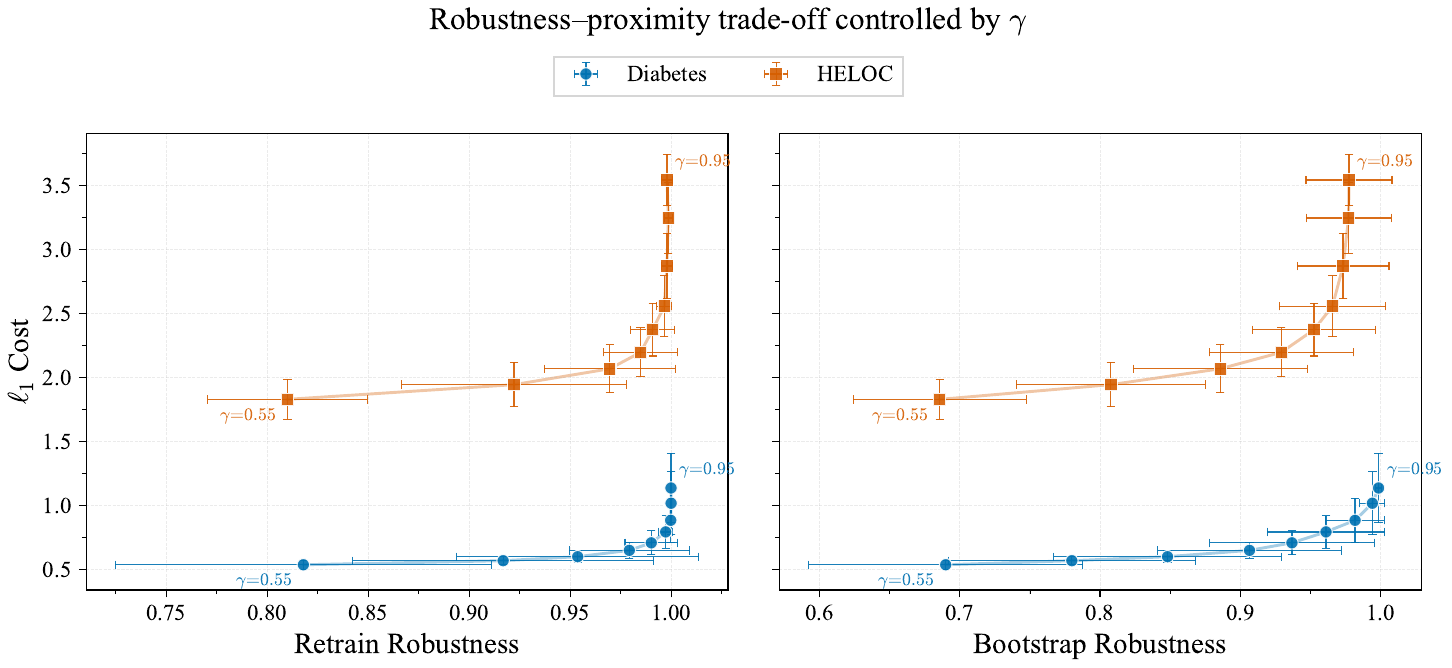}
    \caption{Robustness--proximity tradeoff controlled by $\gamma$ shown for the MLP classifier. Each point corresponds to a different value of $\gamma$ ranging from $0.55$ to $0.95$. The left and right panel show the
    Rob.{\scriptsize(ret)} and Rob.{\scriptsize(bs)}, respectively. Error bars indicate standard deviations across 5 folds.}
    \label{fig:gamma_tradeoff}
\end{figure}

\paragraph{Effect of $\gamma$.}
Figure~\ref{fig:gamma_tradeoff} shows the robustness-proximity trade-off as $\gamma$ varies from $0.55$ to $0.95$. On both datasets, increasing $\gamma$ monotonically improves robustness at a gradual cost in proximity. The trade-off exhibits a clear elbow: moderate values ($\gamma \in [0.7, 0.8]$) capture most of the robustness gain with a modest $L_1$ increase, while the highest values primarily add proximity cost.

\paragraph{Effect of $\alpha$.}
Table~\ref{tab:alpha} shows how the density weight $\alpha$ affects counterfactual quality. Higher $\alpha$ moderately increases $L_1$ cost while slightly improving robustness, as counterfactuals are pushed deeper into high-density consensus regions. However, the effect is small relative to $\gamma$: within each $\gamma$ block, robustness varies by at most $2{-}3$ percentage points across all $\alpha$ values, whereas changing $\gamma$ from $0.7$ to $0.9$ shifts robustness by $3{-}9$ points. Plausibility remains stable across all settings. We fix $\alpha = 5$ as a balanced default in all main experiments.

\begin{table}[ht]
\centering
\caption{Effect of weight $\alpha$ on counterfactual quality (MLP classifier). Mean across 5 folds.}
\label{tab:alpha}
\begin{minipage}[t]{0.48\textwidth}
\centering
\small
\textbf{Diabetes}\\[0.3em]
\begin{tabular}{cccccc}
\toprule
$\gamma$ & $\alpha$ & $L_1$ & Plaus. & Rob.{\scriptsize(ret)} & Rob.{\scriptsize(bs)} \\
\midrule
\multirow{5}{*}{0.70}
 & 1  & 0.52 & 0.243 & 0.964 & 0.871 \\
 & 2  & 0.56 & 0.240 & 0.971 & 0.886 \\
 & 5  & 0.65 & 0.237 & 0.979 & 0.907 \\
 & 10 & 0.73 & 0.236 & 0.980 & 0.916 \\
 & 20 & 0.80 & 0.237 & 0.977 & 0.918 \\
\midrule
\multirow{5}{*}{0.80}
 & 1  & 0.66 & 0.260 & 0.994 & 0.945 \\
 & 2  & 0.70 & 0.258 & 0.996 & 0.953 \\
 & 5  & 0.79 & 0.255 & 0.997 & 0.961 \\
 & 10 & 0.87 & 0.255 & 0.996 & 0.965 \\
 & 20 & 0.94 & 0.256 & 0.997 & 0.969 \\
\midrule
\multirow{5}{*}{0.90}
 & 1  & 0.87 & 0.289 & 1.000 & 0.988 \\
 & 2  & 0.93 & 0.287 & 1.000 & 0.990 \\
 & 5  & 1.02 & 0.285 & 1.000 & 0.994 \\
 & 10 & 1.10 & 0.285 & 1.000 & 0.995 \\
 & 20 & 1.17 & 0.287 & 1.000 & 0.996 \\
\bottomrule
\end{tabular}
\end{minipage}
\hfill
\begin{minipage}[t]{0.48\textwidth}
\centering
\small
\textbf{HELOC}\\[0.3em]
\begin{tabular}{cccccc}
\toprule
$\gamma$ & $\alpha$ & $L_1$ & Plaus. & Rob.{\scriptsize(ret)} & Rob.{\scriptsize(bs)} \\
\midrule
\multirow{5}{*}{0.70}
 & 1  & 1.62 & 0.450 & 0.969 & 0.884 \\
 & 2  & 1.87 & 0.427 & 0.980 & 0.911 \\
 & 5  & 2.19 & 0.409 & 0.985 & 0.929 \\
 & 10 & 2.40 & 0.401 & 0.989 & 0.936 \\
 & 20 & 2.53 & 0.399 & 0.988 & 0.939 \\
\midrule
\multirow{5}{*}{0.80}
 & 1  & 1.99 & 0.463 & 0.991 & 0.947 \\
 & 2  & 2.24 & 0.444 & 0.993 & 0.959 \\
 & 5  & 2.55 & 0.429 & 0.997 & 0.966 \\
 & 10 & 2.76 & 0.426 & 0.997 & 0.969 \\
 & 20 & 2.87 & 0.427 & 0.996 & 0.968 \\
\midrule
\multirow{5}{*}{0.90}
 & 1  & 2.68 & 0.517 & 0.996 & 0.971 \\
 & 2  & 2.94 & 0.505 & 0.998 & 0.975 \\
 & 5  & 3.24 & 0.500 & 0.999 & 0.977 \\
 & 10 & 3.41 & 0.499 & 0.999 & 0.978 \\
 & 20 & 3.52 & 0.502 & 0.999 & 0.979 \\
\bottomrule
\end{tabular}
\end{minipage}
\end{table}

\section{Conclusions}

We proposed a novel probabilistic framework for generating robust counterfactual explanations that jointly models the data distribution and the space of plausible model decisions through a consensus signal derived from a model ensemble. By conditioning a normalizing flow on ensemble agreement levels, our method encodes robustness information once during training and exposes a single interpretable parameter $\gamma$ that controls the robustness level at inference time without retraining. Experimental results on three datasets and two classifier architectures demonstrate that \our{} achieves superior robustness to model change while maintaining competitive proximity, plausibility, and validity.

Our approach has several limitations that suggest directions for future work. First, the quality of the consensus signal depends on the ensemble adequately representing the space of admissible model changes; if the actual model change falls outside the variations captured by the ensemble, the robustness guarantees may not hold. Second, while our method is classifier-agnostic at inference time, it requires training an ensemble and a normalizing flow offline, which may be computationally demanding for very large datasets or high-dimensional feature spaces. Finally, establishing formal probabilistic guarantees on the robustness of the generated counterfactuals, analogous to those provided by BetaRCE but within our generative framework, remains an open and promising direction.

\begin{credits}
\subsubsection{\ackname} 
The research for two first authors was supported by the National Science Centre (Poland) grant No. 2024/55/B/ST6/02100, while the third author was supported by the National Science Centre (Poland) OPUS grant no. 2023/51/B/ST6/00545.

\subsubsection{\discintname}
The authors have no competing interests to declare that are
relevant to the content of this article.
\end{credits}

\bibliographystyle{splncs04}
\bibliography{bib}

\end{document}